\documentclass{article}

\usepackage{arxiv}

\usepackage[utf8]{inputenc} 
\usepackage[T1]{fontenc}    
\usepackage{hyperref}       
\usepackage{url}            
\usepackage{booktabs}       
\usepackage{amsfonts}       
\usepackage{nicefrac}       
\usepackage{microtype}      
\usepackage{xcolor}         
\usepackage{graphicx}
\usepackage{amsthm}
\usepackage{amsmath}

\usepackage{mathtools}
\usepackage{subcaption}
\usepackage{tabularx}
\usepackage{dsfont}
\usepackage{multirow}
\usepackage{wrapfig}
\usepackage{amssymb}
\usepackage{color}
\usepackage{gensymb}
\usepackage{graphicx}
\usepackage{enumitem}

\newcolumntype{C}[1]{>{\centering\arraybackslash}p{#1}}
\usepackage[capitalize,noabbrev]{cleveref}
\usepackage[numbers,sort&compress]{natbib}
\usepackage{siunitx}
\usepackage{booktabs}
\usepackage{siunitx}
\usepackage{algorithm}
\usepackage{algpseudocode}       

\title{From Detrimental to Beneficial: Dynamic Influence-based Valuation and Editing}

\author{%
  Adrian Nyakairu, Hongfu Liu \\
  Department of Computer Science, Brandeis University\\
  \texttt{\{adriannyakairu,hongfuliu\}@brandeis.edu } \\
}

\begin{document}

\maketitle

\begin{abstract}
Data valuation is a cornerstone of data-centric learning, where prior efforts primarily focus on designing algorithms to classify training samples as either beneficial or detrimental for the learning task. However, leveraging these valuation estimates for subsequent data intervention remains underexplored; conventional approaches typically discard or downweight harmful samples, thereby underutilizing available data resources. In this paper, we present Dynamic Influence-based Valuation and Editing (DIVE), a novel and efficient framework that dynamically estimates sample values at the batch level and transforms detrimental data into beneficial contributions. Rather than altering the raw data, DIVE operates at the optimization level by strategically reversing the gradient directions of harmful samples during training, ensuring seamless integration with standard learning procedures with minimal overhead. Extensive empirical evaluations demonstrate that DIVE consistently improves classification performance, maximizes data efficiency, stabilizes optimization, and effectively generalizes to large language model fine-tuning.
\end{abstract}

\section{Introduction}
Recent years have witnessed growing interest in data quality control, giving rise to the field of data-centric learning. Unlike model-centric approaches, which design novel algorithms with fixed training data, data-centric learning focuses on curating data to better suit the learning algorithm and improve performance. In contrast to conventional preprocessing techniques—such as normalization or outlier removal—which are often decoupled from downstream tasks, data-centric learning explicitly evaluates the contribution of data with respect to the target model. This enables more targeted and interpretable interventions across a range of applications, including training subset selection, synthetic data generation, noisy label detection, active learning, and fairness improvement.

Data influence estimation, a core component of data-centric learning, can be broadly categorized into two groups~\cite{hammoudeh2024training}. Retraining-based methods evaluate the influence of a sample by retraining the model with and without it and measuring the resulting performance change, including classical leave-one-out approaches~\cite{cook1982residuals} and Shapley value–based methods~\cite{ghorbani2019data,jia2019towards,kwon2022beta,jia2018efficient}. In contrast, gradient-based methods estimate influence without retraining, most notably through influence functions. While avoiding the cost of retraining, these methods introduce a new computational challenge in approximating the inverse Hessian. The seminal work of~\cite{koh2017understanding} addresses this via a Taylor expansion and LiSSA~\cite{agarwal2017second} for efficient approximation. Subsequent work further improves scalability through stochastic recursive methods~\cite{koh2017understanding}, random projections~\cite{schioppa2022scaling}, Kronecker-factored eigendecomposition~\cite{grosse2023studying}, and low-rank approximations~\cite{park2023trak, kwon2023datainf}. More recently, Hessian-free approaches simplify computation by approximating the inverse Hessian with the identity matrix~\cite{charpiat2019input, pruthi2020estimating, yang2024revisit}. Beyond static estimation, dynamic data valuation has recently attracted increasing attention~\cite{wang2024data, yang2025layer}.

\textbf{Contributions}.
While data-centric learning has developed rapidly, most existing work focuses on designing efficient and accurate data valuation estimators, with limited attention to how these estimates are subsequently utilized. In particular, prior methods typically remove or downweight identified detrimental samples, thereby discarding potentially useful information. In contrast, we aim to fully exploit these detrimental samples by transforming them into beneficial contributions, leading to improved learning performance. We summarize our main contributions as follows:
\vspace{-2mm}
\begin{itemize}[wide=10pt, leftmargin=*]
\item \textit{Concept}: To the best of our knowledge, we are the first to transform detrimental samples into beneficial contributions, making every sample count during learning. This perspective improves data efficiency by leveraging all available samples, rather than discarding potentially useful information. It also leads to more stable and effective optimization by mitigating harmful gradient signals during training.

\item \textit{Technique}: We propose Dynamic Influence-based Valuation and Editing (DIVE), a simple and efficient framework that dynamically estimates data value at the batch level and modifies the gradients of detrimental samples to make them beneficial for model updates. Notably, we do not alter the raw data; instead, we operate at the gradient level, enabling seamless integration into standard optimization procedures with minimal overhead.

\item \textit{Evaluation}: We conduct extensive experiments to demonstrate the effectiveness of DIVE, comparing it against a range of data valuation and dynamic batch curation methods. We further provide diverse in-depth analyses and extend our approach to large language model fine-tuning.
\end{itemize}

\section{Related Work}
In this section, we introduce two most related directions, influence functions and dynamic batch curation, and then highlight the key differences between our paper and the literature. 

\textbf{Influence Function}.
Influence functions, a mainstream tool for data valuation, were originally developed in robust statistics~\cite{hampel1974influence, cook1982residuals, martin1986influence} to quantify the sensitivity of model parameters to perturbations in training data. They were introduced to the machine learning community by~\cite{koh2017understanding} and have since been widely used to estimate sample influence with respect to validation performance.

To address the high computational cost of influence functions, several works propose efficient approximations of the inverse Hessian. LiSSA~\cite{koh2017understanding} leverages Hessian–vector products for stochastic approximation, while EKFAC~\cite{grosse2023studying} exploits structured eigenvalue decompositions. DataInf~\cite{kwon2023datainf} further simplifies influence estimation by leveraging a rank-1 structure of the empirical Hessian, and TRAK~\cite{park2023trak} employs random projections to obtain a tractable kernel-based approximation. More recently, Hessian-free influence methods have been proposed, which approximate the inverse Hessian with an identity matrix for scalability~\cite{charpiat2019input, pruthi2020estimating, yang2024revisit, killamsetty2021grad}.

Beyond classical influence formulations, several recent works revisit influence estimation from an optimization-aware and non-convex perspective. SGD-Influence~\cite{hara2019data} tracks the stochastic gradient descent trajectory and estimates the effect of removing a training sample by reusing intermediate iterates, thereby relaxing the local convexity assumption. MoSo~\cite{tan2023data} models data pruning by approximating a moving-one-sample-out update along the optimization path, while Z0-Inf~\cite{kokhlikyan2025z0} develops a zeroth-order estimator that avoids explicit gradients and Hessian computations.

Beyond static estimation, dynamic data valuation has recently attracted increasing attention~\cite{wang2024data,yang2025layer}. Early approaches approximate dynamic influence by aggregating influence estimates across training checkpoints~\cite{pruthi2020estimating,grosse2023studying,park2023trak}. However, such aggregation fails to capture the evolving nature of influence and may cancel out conflicting signals. More recent work directly estimates data value at each training step, enabling truly dynamic valuation~\cite{wang2024data,yang2025layer}.

\textbf{Dynamic Batch Curation}. The dynamic batch curation paradigm is conceptually related to several research directions. Curriculum learning~\cite{bengio2009curriculum,wang2021survey,soviany2022curriculum} introduces a structured training schedule by presenting samples in an easy-to-hard order, mimicking human learning processes.
Another line focuses on adaptive data selection or reweighting during training to improve model performance. For example, reducible holdout loss selection and its variants~\cite{mindermann2022prioritized} prioritize samples based on difficulty and uncertainty, while JEST~\cite{evans2024data} performs joint multimodal selection to improve training efficiency. ACID/ACED~\cite{udandarao2025active} further targets efficient distillation through adaptive data selection.
A related direction adjusts sampling probabilities based on gradient information. GradientIS and GradientMIS~\cite{salaun2025online,salaun2025multiple} reweight samples according to the norm of their output-layer gradients to reduce stochastic gradient variance and accelerate optimization. More broadly, several recent works aim to improve efficiency by pruning less informative samples based on estimated scores~\cite{toneva2018empirical,raju2021accelerating,he2024large,qin2023infobatch}. 

While techniques like Gradient Reversal~\cite{ganin2015unsupervised,ganin2016domain} or Negative Learning~\cite{kim2019nlnl,kim2021joint} also adjust gradient directions for unlearning or robust training, they inherently rely on external, prior knowledge or heuristics to pre-identify noisy/detrimental samples. In contrast, DIVE integrates sample valuation and gradient editing into a unified data-centric framework. It dynamically discovers which samples are detrimental without any explicit prior labels or loss-threshold heuristics, automatically turning harmful gradients into beneficial updates on the fly.

To the best of our knowledge, existing data-centric learning approaches primarily focus on identifying and removing or downweighting detrimental samples. In contrast, our work takes a fundamentally different perspective by transforming detrimental samples into beneficial contributions at the optimization level. At the technical level, since data valuation evolves during model optimization, we adopt a dynamic formulation that estimates data valuation at each batch. Based on this, we edit detrimental samples at the gradient level, rather than modifying the raw data, enabling seamless integration into standard optimization procedures.

\section{Preliminaries on Influence Functions}\label{sec:pre}
Consider a classifier with parameters $\theta$$\in$$\mathbb{R}^{D}$ mapping instances $z$$=$$\{x,y\}$ from input space $x$$\in$$\mathcal{X}$ to output space $y$$\in$$\mathcal{Y}$, the model parameters $\Hat{\theta}$$=$$\textup{argmin}_{\theta \in \Theta} \frac{1}{n} \sum_{i=1}^n \ell(z_i; \theta)$ can be obtained by solving the empirical risk minimization problem. If we downweight a training sample $z_j$ by a very small fraction $\epsilon$, the new parameters can be $\Hat{\theta}(z_j; -\epsilon) = \textup{argmin}_{\theta \in \Theta} \frac{1}{n} (\sum_{i=1}^n \ell(z_i; \theta)$$-$$\epsilon \ell(z_j; \theta))$. By evaluating the limit as $\epsilon$ approaches 0, the seminal work of \cite{koh2017understanding} provides an estimation for the influence score associated with the removal of $z_j$ from the training set:\vspace{-1mm}
{\small
\begin{equation}\label{eq:influence}
\begin{aligned}
\mathcal{I}(z_j;\Hat{\theta}) = \sum\nolimits_{z \in \mathcal{V}}\nabla\ell(z; \Hat{\theta})^{\top}\mathbf{H}^{-1}_{\Hat{\theta}}\nabla\ell(z_j; \Hat{\theta}),\vspace{-4mm}
\end{aligned}
\end{equation}}
where $\mathcal{V}$ denotes the validation set (self-influence employs the training set instead), $\nabla\ell(z_j; \Hat{\theta})$ is the gradient of sample $z_j$, and $\mathbf{H}_{\Hat{\theta}}$$=$$\sum_{i=1}^n \nabla^2 \ell(z_i; \Hat{\theta})$ denotes the Hessian matrix.

It is well recognized that influence functions rely on several strong assumptions, including model convexity and access to a fully converged solution. While many studies have extended influence functions to deep models with promising empirical results~\cite{chhabra2024what}, they primarily focus on improving computational efficiency, particularly in approximating the inverse Hessian matrix. Common strategies include stochastic recursive approximation~\cite{koh2017understanding}, random projection~\cite{schioppa2022scaling}, Kronecker-Factored eigendecomposition~\cite{grosse2023studying}, low-rank or rank-1 approximations~\cite{park2023trak,kwon2023datainf}, and simplifying the inverse Hessian as the identity matrix, leading to Hessian-free influence functions~\cite{charpiat2019input,pruthi2020estimating,yang2024revisit}.

\section{Method}
In this section, we introduce our Dynamic Influence-based Valuation and Editing (DIVE), a simple and efficient method that estimates the data valuation in the dynamic batch and modifies the gradient of the detrimental samples to be beneficial ones for classification model updating.   

\textbf{Dynamic Data Valuation}. Influence functions are widely used for data valuation, estimating the impact of individual training samples on a target objective, typically the validation loss. Most existing influence function–based methods compute data values using a converged model and then apply these estimates to guide subsequent actions, such as removing detrimental samples and retraining the model on the modified dataset. However, this procedure has two key limitations. First, it produces static data valuations, as the influence scores are derived from a fixed, converged model. As a result, samples identified as detrimental may not remain harmful once the training set is modified and the model is retrained. Second, the approach requires two rounds of training: one to obtain the converged model for valuation and another to retrain on the filtered dataset. This makes it computationally prohibitive for large-scale datasets and models.

Recently, dynamic data valuation has gained increasing attention~\cite{wang2024data,yang2025layer}. Instead of estimating data value based on a fully converged model, these approaches compute data valuation at the mini-batch level during training and use it to guide optimization. While this paradigm addresses the limitations of static valuation—namely, outdated estimates and the need for retraining—it introduces significant computational overhead due to the frequent evaluation of data values. As a result, only highly efficient approximations, such as Hessian-free influence functions, are practical in large-scale settings.

In this paper, we adopt a dynamic data valuation framework. Let $\mathcal{B} = \{z_j\}_{j=1}^b$ denote a mini-batch of size $b$. The influence of a sample $z_j$ at training step $t$ is defined as:
\begin{equation}\label{eq:influence}
\mathcal{I}(z_j; \theta^{(t)}) = \sum\nolimits_{z \in \mathcal{V}} \nabla \ell(z; \theta^{(t)})^{\top}\cdot \nabla \ell(z_j; \theta^{(t)}),
\end{equation}
where $\theta^{(t)}$ is the model parameter at $t$-th epoch. Eq.~\eqref{eq:influence} can be interpreted as a Hessian-free approximation of influence functions that avoids the strong assumptions required by conventional formulations. Intuitively, model updates driven by training samples that are aligned with the validation objective will tend to reduce validation loss. This alignment is not defined in the original feature space, but in the gradient space, which directly governs parameter updates. As a result, Eq.~\eqref{eq:influence} does not rely on assumptions such as model convexity or access to a fully converged solution. Its effectiveness has been empirically validated in prior work \cite{wang2024data} and ~\cite{yang2025layer}. However, these studies are implemented in PyTorch, which only provides batch-level gradients by default, whereas Eq.~\eqref{eq:influence} requires sample-level gradients. To address this limitation, they introduce approximations such as ghost influence and layer-wise estimation.

In contrast, we implement our method in JAX, which natively supports efficient computation of sample-level gradients, enabling direct and accurate estimation of Eq.~\eqref{eq:influence}. Based on the sign of $\mathcal{I}(z_j; \theta^{(t)})$, the batch $\mathcal{B}$ can be partitioned into $\mathcal{B}{+}$ and $\mathcal{B}{-}$, corresponding to samples with positive and negative influence, respectively, for the following editing.

\textbf{Dynamic Gradient Editing}. Unlike existing dynamic data valuation methods~\cite{wang2024data,yang2025layer}, which discard detrimental samples at the batch level, we instead transform such samples into beneficial ones. As shown in Eq.~\eqref{eq:influence}, the valuation is given by the inner product between gradients, which determines whether a sample contributes positively or negatively to the target objective. For samples with negative influence, we flip the direction of their gradients to convert them into beneficial updates.

Importantly, our approach does not modify the raw data, but directly edits the gradients, allowing seamless integration into standard optimization procedures. The parameter update is defined as:
\begin{equation}
\theta^{(t+1)} = \theta^{(t)} - \eta \left( \sum_{z_j \in \mathcal{B}{+}} \nabla \ell(z_j; \theta^{(t)}) - \sum_{z_j \in \mathcal{B}{-}} \nabla \ell(z_j; \theta^{(t)}) \right),
\end{equation}
where $\eta$ is the learning rate. In this way, every sample contributes constructively to the training process, improving data efficiency in dynamic learning.

\section{Experimental Results}
In this section, we first introduce the experimental setup, including datasets, baselines, and implementation details. We then evaluate the performance of DIVE in terms of both effectiveness and efficiency by comparing it with existing baseline methods, followed by in-depth analyses to better understand its behavior. Finally, we extend DIVE to large-scale LLM fine-tuning.

\subsection{Experimental Setup}


\textbf{Datasets.}
We evaluate our method on six noisy-label image benchmarks. \textit{CIFAR-10N} (50,000 training images, 10 classes) provides three human-annotated noise settings—\textit{aggregate} ($\sim$9\%), \textit{random} ($\sim$17\%), and \textit{worst} ($\sim$40\%)—and \textit{CIFAR-100N} (50,000 images, 100 classes) carries $\sim$40\% label noise~\cite{wei2022learning}. \textit{Animal-10N} (50,000 training / 5,000 test images, 10 classes) contains $\sim$8\% human-annotation noise~\cite{song2019selfie}, and \textit{Food-101N} (75,750 training / 25,250 test images, 101 classes) carries $\sim$20\% intrinsic web-label noise in its training split~\cite{bossard2014food}. The noise level thus ranges from $\sim$8\% to $\sim$40\%. We consider two reference settings: (i)~\emph{with a validation set}, where influence is estimated against a small clean reference (1,000 clean-labeled held-out samples; for Animal-10N and Food-101N, which have no clean training labels, we set aside a class-stratified 1/5 of the clean test split as validation and evaluate on the remaining 4/5), and (ii)~\emph{self-influence}, which requires no validation data and uses the mini-batch mean gradient as the reference. The details of datasets and models can be found in Appendix~\ref{app:data}.

\textbf{Baselines.}
We compare against seven baselines across four categories:
(1)~\textit{Vanilla}: standard SGD;
(2)~\textit{Static influence}: LiSSA~\cite{koh2017understanding} and IP~\cite{yang2024revisit};
(3)~\textit{Curriculum}: SPL~\cite{kumar2010self};
(4)~\textit{Dynamic}: InfoBatch~\cite{qin2023infobatch}, Ghost~\cite{wang2024data},
LAI~\cite{yang2025layer}, our DIVE$_{r}$ (removal variant), and our DIVE.

\textbf{Implementation.}
For all image datasets we use a ResNet-9~\cite{he2016deep} backbone with a three-layer
MLP head; sample influence is scored only on the head using cached penultimate features,
so the convolutional backbone is never differentiated per sample. Models are trained with
SGD (Nesterov momentum 0.9, decoupled weight decay $5{\times}10^{-4}$, batch size 512,
standard augmentation) under a one-cycle learning-rate schedule with peak learning rate 0.1
(linear warm-up over the first 15\% of steps, then annealing to $\sim$0). We train for 30
epochs on CIFAR-10N and CIFAR-100N, 40 on Animal-10N, and 100 on Food-101N, applying gradient
clipping (global norm 1.0) on the higher-resolution Animal-10N and Food-101N.
DIVE$_{r}$ and DIVE use a warm-up phase of 30\% of the training epochs, editing every $2$
mini-batches with last-layer gradients. All experiments are implemented in JAX and executed
on Google Cloud TPU v4-8. We report the best test accuracy as mean $\pm$ standard deviation
over 4 seeds.

\begin{table}[t]
\centering
\caption{Accuracy of data-centric methods on noisy benchmarks with a validation set}\label{tab:val}
\resizebox{0.95\linewidth}{!}{
\begin{tabular}{l|cccccc|c}
\toprule
Method & \textit{CIFAR-10Na} & \textit{CIFAR-10Nw} & \textit{CIFAR-10Nr} & \textit{CIFAR-100N} & \textit{Animal-10N} & \textit{Food-101N} & Avg. \\
\midrule
Vanilla      & 88.94$\pm0.16$ & 78.97$\pm$0.22 & 86.62$\pm$0.28 & 54.41$\pm$0.49 & 82.60$\pm$0.26 & 72.42$\pm$0.14 & 77.33 \\
\midrule
LiSSA~(\citeyear{koh2017understanding})        & 89.03$\pm$0.50 & 78.38$\pm$0.50 & 86.32$\pm$0.13 & 54.67$\pm$0.30 & 82.41$\pm$0.25 & 72.07$\pm$0.22 & 77.15 \\
IP~(\citeyear{yang2024revisit})           & 89.54$\pm$0.19 & 77.82$\pm$0.40 & 86.54$\pm$0.11 & 55.03$\pm$0.18 & 82.72$\pm$0.35 & 72.13$\pm$0.27 & 77.30 \\
\midrule
SPL~(\citeyear{kumar2010self})          & 88.98$\pm$0.26 & 70.53$\pm$0.85 & 86.32$\pm$0.34 & 38.47$\pm$0.55 & 83.38$\pm$0.18 & \textbf{72.97$\pm$0.07} & 73.44 \\
\midrule
InfoBatch~(\citeyear{qin2023infobatch})    & 89.64$\pm$0.17 & 80.90$\pm$0.34 & 87.92$\pm$0.21 & 56.36$\pm$0.47 & 83.52$\pm$0.30 & 72.83$\pm$0.09 & 78.53 \\
GHOST~(\citeyear{wang2024data})        & \textbf{90.11$\pm$0.16} & 83.25$\pm$0.26 & 88.74$\pm$0.17 & \textbf{57.52$\pm$0.43} & 84.10$\pm$0.38 & 72.73$\pm$0.16 & 79.41 \\
LAI~(\citeyear{yang2025layer})          & 90.10$\pm$0.12 & 82.98$\pm$0.25 & 88.63$\pm$0.13 & \textbf{57.52$\pm$0.47} & 84.32$\pm$0.19 & 72.64$\pm$0.20 & 79.36 \\
DIVE$_{r}$ (Ours) & 89.69$\pm$0.05 & 81.54$\pm$0.31 & 88.11$\pm$0.16 & 56.84$\pm$0.43 & 83.44$\pm$0.25 & 72.59$\pm$0.18 & 78.70 \\
DIVE (Ours) & {90.00$\pm$0.16} & \textbf{83.48$\pm$0.18} & \textbf{88.77$\pm$0.22} & {57.38$\pm$0.52} & \textbf{84.53$\pm$0.17} & {72.91$\pm$0.26} & \textbf{79.51} \\
\bottomrule
\end{tabular}
}
\end{table}

\begin{table}
\centering
\caption{Training overhead relative to the vanilla baseline.}
\label{tab:overhead}
\resizebox{0.88\linewidth}{!}{
\begin{tabular}{l|cccccc}
\toprule
Method & \textit{CIFAR-10Na} & \textit{CIFAR-10Nw} & \textit{CIFAR-10Nr} & \textit{CIFAR-100N} & \textit{Animal-10N} & \textit{Food-101N} \\
\midrule
Vanilla & 1.00 & 1.00 & 1.00 & 1.00 & 1.00 & 1.00 \\
\midrule
LiSSA~(\citeyear{koh2017understanding}) & 2.50 & 2.65 & 2.64 & 2.62 & 2.80 & 2.50 \\
IP~(\citeyear{yang2024revisit}) & 2.50 & 2.64 & 2.62 & 2.65 & 2.80 & 2.51 \\
\midrule
SPL~(\citeyear{kumar2010self}) & 1.00 & 1.00 & 1.00 & 1.01 & 1.01 & 1.00 \\
\midrule
InfoBatch~(\citeyear{qin2023infobatch}) & 1.01 & 1.02 & 1.02 & 1.01 & 1.02 & 1.01 \\
GHOST~(\citeyear{wang2024data}) & 1.22 & 1.22 & 1.22 & 1.22 & 1.25 & 1.22 \\
LAI~(\citeyear{yang2025layer}) & 1.18 & 1.18 & 1.19 & 1.19 & 1.24 & 1.21 \\
DIVE$_r$ (Ours) & 1.09 & 1.09 & 1.09 & 1.09 & 1.12 & 1.11 \\
DIVE (Ours) & 1.09 & 1.09 & 1.09 & 1.09 & 1.12 & 1.11 \\
\bottomrule
\end{tabular}
}
\vspace{-3mm}
\end{table}

\subsection{Algorithmic Performance}
To evaluate the effectiveness of DIVE, we consider the task of classification with noisy labels. Table~\ref{tab:val} reports the average classification accuracy and standard deviation over five runs on six noisy benchmark datasets. Overall, DIVE consistently achieves the best performance on three out of six datasets and attains the highest average accuracy of 0.7951, demonstrating strong robustness to label noise across diverse tasks. Notably, DIVE significantly outperforms the second-best method in terms of average performance under a paired t-test ($p < 0.1$).

Compared with static influence-based methods (e.g., LiSSA and IP), DIVE consistently achieves substantial improvements across all datasets, demonstrating the benefits of dynamic influence estimation over static approximations. DIVE also consistently outperforms curriculum-based learning (SPL), suggesting that simple curriculum heuristics are ineffective under severe label noise. In fact, SPL even underperforms the vanilla baseline. Among dynamic approaches, InfoBatch, Ghost, and LAI are all competitive and consistently outperform static methods; nevertheless, DIVE further achieves consistent gains over these strong baselines. 

Beyond the above high-level observations, several additional insights can be drawn. (1) Ghost, LAI, DIVE$_r$, and DIVE all belong to the same dynamic data curation framework and employ the same first-order influence estimation, differing only in how identified detrimental samples are handled. Ghost and LAI remove detrimental samples at every training batch, whereas DIVE$_r$ performs removal less frequently (once every second mini-batch). This less aggressive intervention substantially reduces the computational overhead while maintaining competitive performance, suggesting that overly frequent data curation is unnecessary. (2) DIVE further extends DIVE$_r$ by transforming detrimental samples into beneficial ones instead of simply removing them. As shown in Table~\ref{tab:val}, DIVE consistently outperforms DIVE$_r$ across all datasets, with statistically significant improvements. This suggests that transforming detrimental samples better preserves informative yet noisy training examples, allowing the model to exploit their underlying signal rather than discarding potentially useful information. Consequently, DIVE achieves more stable optimization and better generalization under noisy supervision.

To complement the above analysis, we further examine the computational efficiency of different data curation methods. Table~\ref{tab:overhead} reports the training overhead relative to the vanilla baseline. LiSSA and IP require two rounds of training and therefore incur the highest computational overhead. SPL follows a pre-defined curriculum and incurs only marginal runtime overhead. Among dynamic methods, InfoBatch performs data curation at every batch while remaining relatively efficient, as it relies only on forward losses without computing sample-level gradients. However, as shown in Table~\ref{tab:val}, this efficiency comes at the cost of inferior accuracy compared with gradient-based approaches. Ghost, LAI, DIVE$_r$, and DIVE all rely on sample-level gradients. Compared with Ghost and LAI, DIVE incurs lower computational overhead by performing detrimental sample editing once every other epoch instead of at every mini-batch. This less frequent intervention not only substantially improves efficiency but also avoids excessive interference with the optimization process, leading to better overall performance. Finally, although DIVE further transforms detrimental samples into beneficial ones, this operation introduces virtually no additional computational overhead compared with DIVE$_r$.

\subsection{In-depth Exploration}

\begin{figure}[t]
\centering
\includegraphics[width=\linewidth]{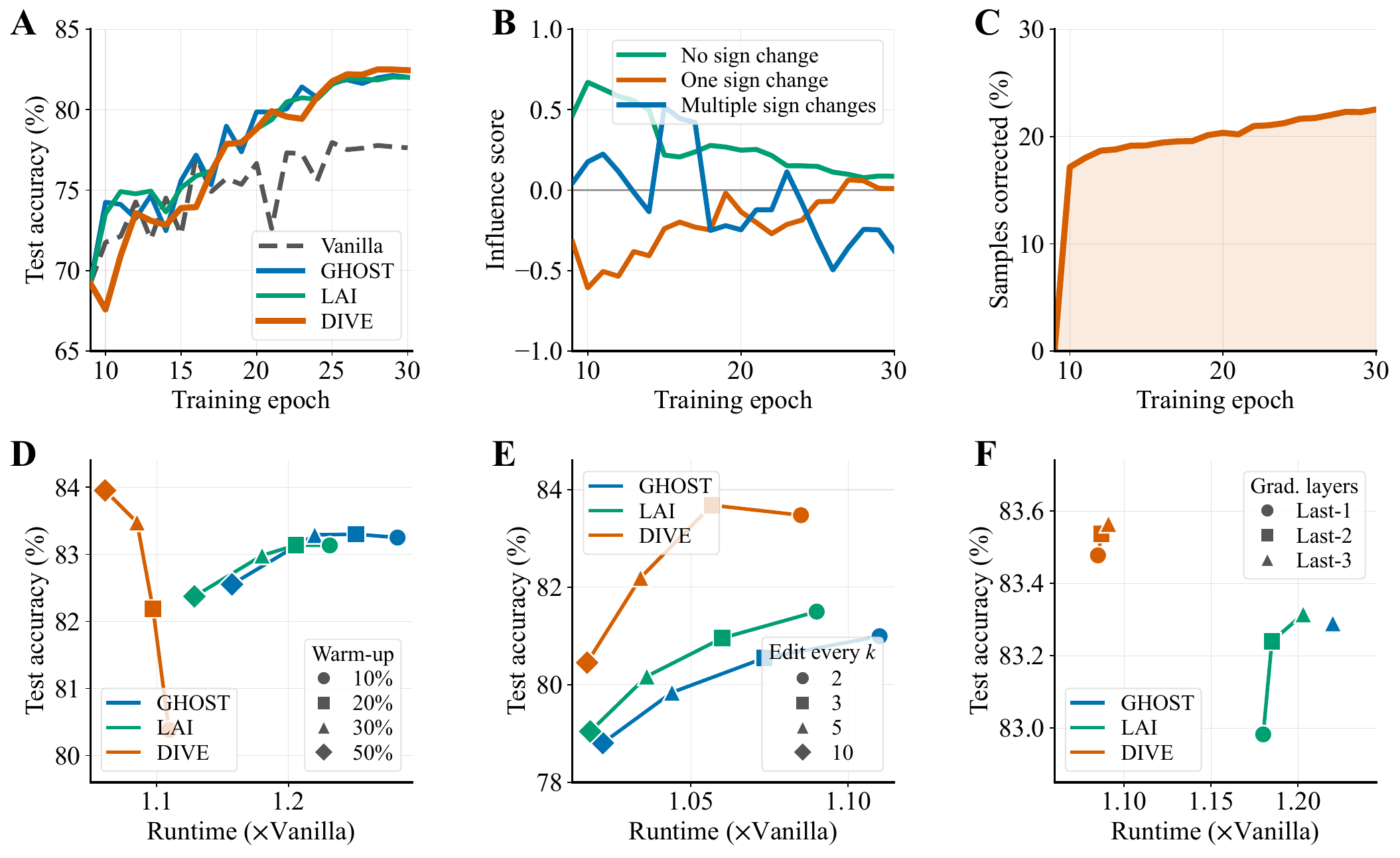}
\caption{In-depth exploration of DIVE on \textit{CIFAR-10N-w}.
\textbf{(A)} Test accuracy vs.\ epoch, shown from the end of warm-up.
\textbf{(B)} Influence scores of three individual training samples across epochs (from the end of warm-up). 
\textbf{(C)} Fraction of samples corrected by DIVE per epoch.
In (D)--(F) the horizontal axis is training runtime relative to vanilla, colour denotes the method, and marker shape denotes the ablated hyperparameter; points of the same method are joined by a line.
\textbf{(D)} Warm-up-duration ablation (marker: 10/20/30/50\%).
\textbf{(E)} Editing-frequency ablation (marker: edit every $k{\in}\{2,3,5,10\}$ mini-batches).
\textbf{(F)} Scoring-layer ablation (marker: last 1/2/3 gradients for DIVE and LAI; GHOST uses the full head, a single point). }\vspace{-8mm}
\label{fig:indepth}
\end{figure}

Beyond reporting the algorithmic performance in terms of effectiveness and efficiency in the above, we continue to provide diverse in-depth explorations of our DIVE as follows.

\textbf{Detrimental Sample Verification}. In the data valuation literature, sample valuation is typically evaluated by measuring the correlation between the estimated values and the leave-one-out loss changes~\cite{koh2017understanding,wang2024data}. In contrast, our work focuses on identifying detrimental samples, rather than precisely estimating their exact values. Although the influence estimation in Eq.~\eqref{eq:influence} provides only a coarse approximation, it still preserves the polarity of samples—i.e., whether they are beneficial or detrimental. Based on this observation, we adopt a more direct evaluation criterion: a sample is considered correctly identified as detrimental if correcting it leads to an improvement in model performance. This criterion better aligns with our goal of effective detrimental sample identification, rather than accurate correction. 

Figure~\ref{fig:indepth}A illustrates the test performance of several dynamic data curation methods and the vanilla baseline on \textit{CIFAR-10N-w} across training epochs. After the warm-up phase, DIVE consistently outperforms the vanilla method, with the performance gains primarily attributed to the correction of detrimental samples. Beyond verifying the effectiveness of detrimental sample correction, two additional observations can be made.
(1) During the warm-up phase, the vanilla method achieves the best performance and exhibits a steady upward trend, suggesting that sample quantity is the dominant factor at this stage. As training progresses, the dominant factor shifts to sample quality, where improvements are achieved by either removing or correcting detrimental samples.
(2) Correcting detrimental samples is a stronger operation than removing them: as reported in Table~\ref{tab:self}, DIVE$_r$ (removal) already improves over the vanilla baseline, yet DIVE attains a consistently higher accuracy (e.g., $82.67$ vs.\ $80.98$ on \textit{CIFAR-10N-w} and $79.03$ vs.\ $78.17$ on average), demonstrating that the gains stem more from \emph{correcting} detrimental samples than from merely discarding them.

\textbf{Dynamic Mechanism}. We further examine how DIVE acts on the training set as optimization proceeds. Figure~\ref{fig:indepth}B tracks the influence score of three representative training examples across epochs and shows that a sample's influence \emph{sign} is not constant: the same sample can be beneficial (score above zero) at some epochs and detrimental (below zero) at others. The three examples differ only in how often this sign flips---one never changes sign, remaining beneficial throughout; a second changes sign once, starting detrimental and becoming beneficial as training proceeds; and a third changes sign multiple times, oscillating between beneficial and detrimental. Whether a sample helps or harms the model is therefore not an intrinsic, fixed property but depends on the current state of optimization. Figure~\ref{fig:indepth}C aggregates this effect over the whole training set, reporting the fraction of samples corrected by DIVE at each epoch: no editing occurs during the warm-up phase, after which the correction rate rises as the model fits the data and stabilizes at around 20\%, remaining consistently non-zero for the remainder of training. Together these confirm that the set of detrimental samples is not static but continuously evolves as optimization proceeds, and thus an adaptive, per-step correction strategy is necessary rather than a one-shot static valuation.

\textbf{Performance-Cost Analysis}. We analyze the performance-cost of DIVE from three perspectives: the warm-up duration, the choice of gradient computation across different layers, and the frequency of detrimental sample correction.

\textit{Warm-up Duration}. Figure~\ref{fig:indepth}D plots accuracy against runtime as the warm-up duration varies; a longer warm-up shortens the dynamic editing phase and therefore also lowers runtime. DIVE's accuracy keeps improving as warm-up lengthens (from $10\%$ to $50\%$ of epochs)---becoming both more accurate and cheaper---reflecting that a better-fit reference model yields more reliable influence estimates; we nonetheless adopt a $30\%$ warm-up as the default, which already captures most of the gain and matches the budget used across all main experiments. In contrast, the removal-based GHOST/LAI peak early and then decline, since prolonged vanilla training leaves fewer clearly-detrimental samples for them to profitably discard.

\textit{Editing Frequency and Scoring Layers}. We further study the frequency of detrimental sample editing, defined as editing once every $k$ mini-batches during the dynamic phase; a larger $k$ reduces the editing runtime. As shown in Figure~\ref{fig:indepth}E, DIVE's accuracy peaks at a small $k$ ($k{=}2$ and $3$) and then declines as $k$ grows, yet it dominates GHOST and LAI at every runtime budget. Very frequent editing applies gradient ascent to the detrimental samples too aggressively---at the extreme of every mini-batch it destabilizes training and causes a sharp accuracy drop---whereas very infrequent editing (e.g., $k{=}10$) fails to correct the accumulated noise in time. A moderate frequency ($k{=}2$) achieves the best accuracy--runtime balance, which we adopt as the default. In contrast, the removal-based methods (GHOST, LAI) exhibit the \emph{opposite} accuracy trend and monotonically favor more frequent editing, as discarding samples is gentler and does not destabilize optimization. 

\textit{Scoring-layer ablation}. Figure~\ref{fig:indepth}F ablates the number of gradient layers used for scoring: DIVE attains the highest accuracy at the lowest runtime and is insensitive to the number of last-$k$ gradient layers, whereas LAI must include more gradient layers---raising its cost toward GHOST, which always scores over the full head---to improve. This indicates that DIVE's sign-correction remains effective even from a single last-layer gradient, keeping its scoring cost minimal.

Based on these observations, we use the last-layer gradient, a 30\% warm-up duration, and editing every $2$ mini-batches ($k{=}2$) as the default configuration. Similar phenomena on \textit{Animal-10N} and \textit{Food-101N} are shown in Appendix~\ref{app:experiment}.

\begin{table}[t]
\centering
\caption{Accuracy of data-centric methods on noisy benchmarks without a validation set}\label{tab:self}
\resizebox{0.95\linewidth}{!}{
\begin{tabular}{l|cccccc|c}
\toprule
Method & \textit{CIFAR-10Na} & \textit{CIFAR-10Nw} & \textit{CIFAR-10Nr} & \textit{CIFAR-100N} & \textit{Animal-10N} & \textit{Food-101N} & Avg. \\
\midrule
Vanilla      & 88.94$\pm$0.16 & 78.97$\pm$0.22 & 86.62$\pm$0.28 & 54.41$\pm$0.49 & 82.60$\pm$0.26 & 72.42$\pm$0.14 & 77.33 \\
\midrule
LiSSA~(\citeyear{koh2017understanding}) & 88.73$\pm$0.15 & 78.38$\pm$0.50 & 86.23$\pm$0.22 & 54.45$\pm$0.15 & 82.41$\pm$0.25 & 72.00$\pm$0.10 & 77.03 \\
IP~(\citeyear{yang2024revisit}) & 88.67$\pm$0.53 & 78.31$\pm$0.42 & 86.56$\pm$0.49 & 54.55$\pm$0.25 & 82.60$\pm$0.45 & 72.03$\pm$0.16 & 77.12 \\
\midrule
SPL~(\citeyear{kumar2010self}) & 89.00$\pm$0.17 & 70.53$\pm$0.85 & 86.01$\pm$0.45 & 38.47$\pm$0.55 & 83.27$\pm$0.51 & 72.79$\pm$0.13 & 73.34 \\
\midrule
InfoBatch~(\citeyear{qin2023infobatch}) & 89.74$\pm$0.06 & 81.12$\pm$0.24 & 88.02$\pm$0.19 & 56.42$\pm$0.35 & 83.29$\pm$0.40 & \textbf{72.80$\pm$0.22} & 78.56 \\
GHOST~(\citeyear{wang2024data}) & 89.71$\pm$0.09 & 82.26$\pm$0.32 & 88.45$\pm$0.17 & \textbf{55.96$\pm$0.30} & 84.16$\pm$0.19 & 72.70$\pm$0.09 & 78.87 \\
LAI~(\citeyear{yang2025layer}) & 89.77$\pm$0.09 & 82.34$\pm$0.42 & 88.42$\pm$0.06 & 55.63$\pm$0.39 & 84.10$\pm$0.28 & 72.50$\pm$0.22 & 78.79 \\
DIVE$_{r}$ (Ours) & 89.52$\pm$0.11 & 80.98$\pm$0.33 & 87.78$\pm$0.18 & 55.02$\pm$0.59 & 83.13$\pm$0.25 & 72.57$\pm$0.21 & 78.17 \\
DIVE (Ours) & \textbf{90.00$\pm$0.16} & \textbf{82.67$\pm$0.17} & \textbf{88.64$\pm$0.12} & 55.66$\pm$0.37 & \textbf{84.61$\pm$0.27} & 72.61$\pm$0.20 & \textbf{79.03} \\
\bottomrule
\end{tabular}
}\vspace{-4mm}
\end{table}

\textbf{Self-Influence}. We further evaluate DIVE in the absence of a validation set by adopting self-influence, where training samples are used to estimate influence scores. Specifically, the first term in Eq.~\eqref{eq:influence} is replaced by the average gradient of the training samples within each mini-batch. As shown in Table~\ref{tab:self}, DIVE consistently achieves the best average performance and significantly outperforms the second-best method ($p<0.1$), demonstrating that our method remains effective even when no clean validation data are available.

\subsection{Extension on Fine-Tuning LLMs}

Beyond the vision benchmarks considered above, we further examine whether DIVE remains effective in the setting of LLM fine-tuning, where the base model is orders of magnitude larger, training is limited to a single epoch, and label noise is inherent to the datasets rather than synthetically injected. This regime differs from the image benchmarks in three important aspects. First, per-sample full-model gradients are infeasible—Gemma-3-12B has $\sim$12B parameters, so per-sample gradient tensors on a mini-batch would exceed device memory by orders of magnitude. Following the last-layer approximation identified in the performance-cost analysis, we restrict influence estimation to the LoRA adapters only. Second, training is short (one epoch) with a cosine-decayed learning-rate schedule and warm-up, so an aggressive intervention schedule is not appropriate; DIVE is invoked only after a fraction of the epoch is completed. Third, a validation split is not always available with a clean, distribution-matched reference (e.g., \textit{Civil\_Comments} follows a natural $\sim$8/92 toxic/non-toxic distribution), so we adopt the self-influence variant developed in the previous section, using the mean gradient of the current training mini-batch as the reference direction.

\begin{table*}[t]
\centering
\caption{Performance of DIVE for LLM fine-tuning}\label{tab:performance_llm}
\resizebox{0.75\textwidth}{!}{
\begin{tabular}{l |ccc|cccc}
\toprule
\multirow{2}{*}{Dataset}  & \multirow{2}{*}{\#Class} & \multirow{2}{*}{\#Train} & \multirow{2}{*}{\#Test} & \multicolumn{2}{c}{Test Loss} & \multicolumn{2}{c}{Test Accuracy} \\
\cmidrule(lr){5-6}
\cmidrule(lr){7-8}
 & &  &  & Vanilla & DIVE & Vanilla & DIVE \\
\midrule
\textit{PAWS} & 2 & 49,401 & 8,000 & 0.1584 & \textbf{0.1549} & \textbf{0.9515} & 0.9514\\
\textit{ANLI\_r3}  & 3 & 100,459 & 1,200 &   0.8049 & \textbf{0.7374} & 0.6806 & \textbf{0.6988}   \\
\textit{Civil\_Comments\_b} & 2 & 200,000 & 97,320 &  0.1366 & \textbf{0.1150 }& 0.8377  & \textbf{0.8590}  \\
\textit{Civil\_Comments} & 2 & 1,804,874 & 97,320 &  0.0702 & \textbf{0.0676} & \textbf{0.9531 } & 0.9518\\
\bottomrule
\end{tabular}
}\vspace{-4mm}
\end{table*}

\textbf{Experimental Setting.} We fine-tune \textit{Gemma-3-12B-IT} in a generative supervised-fine-tuning formulation: each classification example is verbalized as a prompt–response pair, and the loss is computed only on the response tokens. LoRA adapters (rank 16, $\alpha=32$) are attached to all attention and MLP projections; all other parameters are frozen. Optimization uses Adam with peak learning rate $2\!\times\!10^{-4}$, a linear warm-up over the first few hundred steps followed by cosine decay to zero, effective batch size 16, and one epoch over the training set. The DIVE schedule mirrors the CV configuration: the first 30\% of the epoch runs vanilla SFT as warm-up, after which DIVE editing and vanilla optimization alternate in a repeating cycle of $1\%$ DIVE bursts followed by $9\%$ vanilla bursts. Evaluation is greedy decoding on the full test split of each dataset, and predictions are matched against the canonical label vocabulary via exact string match. All runs are executed on a Google Cloud TPU v5p-8.

We consider four text classification datasets that span sentence-pair semantic reasoning (\textit{PAWS}), adversarially collected natural-language inference (\textit{ANLI\_r3}), and toxicity detection under both balanced and natural class distributions (\textit{Civil\_Comments\_b}: 200K stratified 50/50; \textit{Civil\_Comments}: the full 1.8M natural distribution with $\sim$8\% toxic). These four datasets differ substantially in scale (49K–1.8M training examples), number of classes (2–3), and difficulty, allowing us to probe DIVE's behavior across a range of LLM fine-tuning regimes.

\textbf{Results and Analysis.} Table~\ref{tab:performance_llm} reports test loss and test accuracy of vanilla SFT and DIVE. \emph{DIVE achieves lower test loss than vanilla SFT on every dataset without exception}—the reduction ranges from 2.2\% (\textit{PAWS}: 0.1584$\to$0.1549) and 3.7\% (\textit{Civil\_Comments}: 0.0702$\to$0.0676) on the two saturated datasets to 8.4\% (\textit{ANLI\_r3}: 0.8049$\to$0.7374) and 15.8\% (\textit{Civil\_Comments\_b}: 0.1366$\to$0.1150) on the two datasets that leave the model meaningful headroom. This uniform reduction in test loss shows that correcting the sign of detrimental samples yields better-calibrated predictions on held-out data across the full range of LLM fine-tuning regimes we consider.

On accuracy, DIVE delivers substantial gains on the two datasets where the vanilla baseline has meaningful headroom—\textit{ANLI\_r3} (+1.82 points, from 68.06 to 69.88) and \textit{Civil\_Comments\_b} (+2.13 points, from 83.77 to 85.90)—while remaining essentially on par on \textit{PAWS} and \textit{Civil\_Comments}, where vanilla SFT already reaches $\sim$95\% and the room for improvement is small. The magnitude of the accuracy gain correlates with the noise level and task difficulty implicit in each dataset: \textit{ANLI\_r3} is adversarially curated so its training set contains many hard or ambiguous examples whose per-sample gradients frequently disagree with the batch consensus, and \textit{Civil\_Comments\_b} contains inherent human-annotation noise (toxicity thresholded at 0.5 without adjudication). In contrast, \textit{PAWS} and full \textit{Civil\_Comments} are cleaner or dominated by an easy majority class, leaving fewer detrimental samples for DIVE to correct. Crucially, the uniform improvement in test loss coupled with preserved accuracy on the saturated datasets shows that DIVE's editing never degrades the model in the easy regime: the sign flip is triggered only for samples whose gradient direction actually disagrees with the batch mean, so on well-fit data the intervention rate is naturally low. These findings confirm that DIVE is directly applicable to large-model text fine-tuning, with no algorithmic modification other than restricting influence computation to the trainable LoRA subspace.

\section{Conclusion}
In this paper, we proposed Dynamic Influence-based Valuation and Editing ({DIVE}), a lightweight and dynamic data curation framework designed to handle noisy and detrimental data during training. Unlike existing methods that primarily focus on identifying and removing detrimental samples, DIVE goes a step further by transforming detrimental samples into beneficial ones, effectively recovering informative underlying signals rather than discarding them. Through extensive experiments across standard image benchmarks and LLM fine-tuning regimes, DIVE consistently demonstrates superior performance and computational efficiency, remaining highly effective even in the absence of clean validation data via self-influence. Our findings highlight the importance of dynamic data valuation and open up promising avenues for repurposing, rather than discarding, noisy training data in large-scale model optimization.

\section*{Acknowledgment}
We gratefully acknowledge the support of the Google TPU Builders Program and Google Tunix library, which provided support and access to the computational framework. We express our gratitude to Dr. Han Yue for assistance with debugging the code.

\bibliographystyle{unsrtnat}   
\bibliography{ref}

@inproceedings{koh2017understanding,
  title={Understanding black-box predictions via influence functions},
  author={Koh, Pang Wei and Liang, Percy},
  booktitle={International Conference on Machine Learning},
  year={2017},
  organization={}
}

@article{park2023trak,
  title={Trak: Attributing model behavior at scale},
  author={Park, Sung Min and Georgiev, Kristian and Ilyas, Andrew and Leclerc, Guillaume and Madry, Aleksander},
  journal={arXiv preprint arXiv:2303.14186},
  year={2023}
}

@article{wang2024data,
  title={Data Shapley in One Training Run},
  author={Wang, Jiachen T and Mittal, Prateek and Song, Dawn and Jia, Ruoxi},
  journal={arXiv preprint arXiv:2406.11011},
  year={2024}
}

@inproceedings{mindermann2022prioritized,
  title={Prioritized training on points that are learnable, worth learning, and not yet learnt},
  author={Mindermann, S{\"o}ren and Brauner, Jan M and Razzak, Muhammed T and Sharma, Mrinank and Kirsch, Andreas and Xu, Winnie and H{\"o}ltgen, Benedikt and Gomez, Aidan N and Morisot, Adrien and Farquhar, Sebastian and others},
  booktitle={International Conference on Machine Learning},
  year={2022}
}

@inproceedings{killamsetty2021grad,
  title={Grad-match: Gradient matching based data subset selection for efficient deep model training},
  author={Killamsetty, Krishnateja and Durga, Sivasubramanian and Ramakrishnan, Ganesh and De, Abir and Iyer, Rishabh},
  booktitle={International Conference on Machine Learning},
  year={2021}
}

@article{kumar2010self,
  title={Self-paced learning for latent variable models},
  author={Kumar, M and Packer, Benjamin and Koller, Daphne},
  journal={Advances in Neural Information Processing Systems},
  year={2010}
}

@article{soviany2022curriculum,
  title={Curriculum learning: A survey},
  author={Soviany, Petru and Ionescu, Radu Tudor and Rota, Paolo and Sebe, Nicu},
  journal={International Journal of Computer Vision},
  volume={130},
  number={6},
  pages={1526--1565},
  year={2022},
  publisher={Springer}
}

@article{wang2021survey,
  title={A survey on curriculum learning},
  author={Wang, Xin and Chen, Yudong and Zhu, Wenwu},
  journal={IEEE Transactions on Pattern Analysis and Machine Intelligence},
  volume={44},
  number={9},
  pages={4555--4576},
  year={2021},
  publisher={IEEE}
}

@inproceedings{bengio2009curriculum,
  title={Curriculum learning},
  author={Bengio, Yoshua and Louradour, J{\'e}r{\^o}me and Collobert, Ronan and Weston, Jason},
  booktitle={International Conference on Machine Learning},
  year={2009}
}

@article{yang2024revisit,
  title={Revisit, Extend, and Enhance Hessian-Free Influence Functions},
  author={Yang, Ziao and Yue, Han and Chen, Jian and Liu, Hongfu},
  journal={Transactions on Machine Learning Research},
  year={2026}
}

@article{martin1986influence,
  title={Influence functionals for time series},
  author={Martin, R Douglas and Yohai, Victor J},
  journal={The Annals of Statistics},
  year={1986},
  publisher={JSTOR}
}

@article{charpiat2019input,
  title={Input similarity from the neural network perspective},
  author={Charpiat, Guillaume and Girard, Nicolas and Felardos, Loris and Tarabalka, Yuliya},
  journal={Advances in Neural Information Processing Systems},
  year={2019}
}

@book{cook1982residuals,
  title={Residuals and influence in regression},
  author={Cook, R Dennis and Weisberg, Sanford},
  year={1982},
  publisher={New York: Chapman and Hall}
}

@article{hampel1974influence,
  title={The influence curve and its role in robust estimation},
  author={Hampel, Frank R},
  journal={Journal of the American Statistical Association},
  year={1974},
  publisher={Taylor \& Francis}
}

@inproceedings{kim2021joint,
  title={Joint negative and positive learning for noisy labels},
  author={Kim, Youngdong and Yun, Juseung and Shon, Hyounguk and Kim, Junmo},
  booktitle={IEEE/CVF Conference on Computer Vision and Pattern Recognition},
  year={2021}
}

@inproceedings{kim2019nlnl,
  title={Nlnl: Negative learning for noisy labels},
  author={Kim, Youngdong and Yim, Junho and Yun, Juseung and Kim, Junmo},
  booktitle={2019 IEEE/CVF International Conference on Computer Vision},
  year={2019}
}

@article{ganin2016domain,
  title={Domain-adversarial training of neural networks},
  author={Ganin, Yaroslav and Ustinova, Evgeniya and Ajakan, Hana and Germain, Pascal and Larochelle, Hugo and Laviolette, Fran{\c{c}}ois and March, Mario and Lempitsky, Victor},
  journal={Journal of Machine Learning Research},
  volume={17},
  number={59},
  pages={1--35},
  year={2016}
}

@inproceedings{ganin2015unsupervised,
  title={Unsupervised domain adaptation by backpropagation},
  author={Ganin, Yaroslav and Lempitsky, Victor},
  booktitle={International Conference on Machine Learning},
  year={2015}
}

@inproceedings{jia2019towards,
  title={Towards efficient data valuation based on the {S}hapley value},
  author={Jia, Ruoxi and Dao, David and Wang, Boxin and Hubis, Frances Ann and Hynes, Nick and G{\"u}rel, Nezihe Merve and Li, Bo and Zhang, Ce and Song, Dawn and Spanos, Costas J},
  booktitle={International Conference on Artificial Intelligence and Statistics},
  year={2019},
  organization={}
}

@inproceedings{ghorbani2019data,
  title={Data shapley: Equitable valuation of data for machine learning},
  author={Ghorbani, Amirata and Zou, James},
  booktitle={International Conference on Machine Learning},
  year={2019},
  organization={}
}

@article{hara2019data,
  title={Data cleansing for models trained with SGD},
  author={Hara, Satoshi and Nitanda, Atsushi and Maehara, Takanori},
  journal={Advances in Neural Information Processing Systems},
  year={2019}
}

@article{tan2023data,
  title={Data pruning via moving-one-sample-out},
  author={Tan, Haoru and Wu, Sitong and Du, Fei and Chen, Yukang and Wang, Zhibin and Wang, Fan and Qi, Xiaojuan},
  journal={Advances in Neural Information Processing Systems},
  year={2023}
}

@article{kokhlikyan2025z0,
  title={Z0-Inf: Zeroth Order Approximation for Data Influence},
  author={Kokhlikyan, Narine and Chaudhuri, Kamalika and Mahloujifar, Saeed},
  journal={arXiv preprint arXiv:2510.11832},
  year={2025}
}

@inproceedings{salaun2025online,
  title={Online Importance Sampling for Stochastic Gradient Optimization},
  author={Sala{\"u}n, Corentin and Huang, Xingchang and Georgiev, Iliyan and Mitra, Niloy J and Singh, Gurprit},
  booktitle={International Conference on Pattern Recognition Applications and Methods},
  year={2025}
}

@inproceedings{salaun2025multiple,
  title={Multiple Importance Sampling for Stochastic Gradient Optimization},
  author={Sala{\"u}n, Corentin and Huang, Xingchang and Georgiev, Iliyan and Mitra, Niloy J and Singh, Gurprit},
  booktitle={International Conference on Pattern Recognition Applications and Methods},
  year={2025}
}

@inproceedings{kwon2022beta,
  title={Beta {S}hapley: a {U}nified and {N}oise-reduced {D}ata {V}aluation {F}ramework for {M}achine {L}earning},
  author={Kwon, Yongchan and Zou, James},
  booktitle={International Conference on Artificial Intelligence and Statistics},
  year={2022},
  organization={}
}

@inproceedings{schioppa2022scaling,
  title={Scaling up influence functions},
  author={Schioppa, Andrea and Zablotskaia, Polina and Vilar, David and Sokolov, Artem},
  booktitle={AAAI Conference on Artificial Intelligence},
  year={2022}
}

@article{jia2018efficient,
  title={Efficient task specific data valuation for nearest neighbor algorithms},
  author={Jia, Ruoxi and Dao, David and Wang, Boxin and Hubis, Frances Ann and Gurel, Nezihe Merve and Li, Bo and Zhang, Ce and Spanos, Costas and Song, Dawn},
  journal={In International Conference on Very Large Data Bases Endowment},
  year={2018},
  }

@article{pruthi2020estimating,
  title={Estimating training data influence by tracing gradient descent},
  author={Pruthi, Garima and Liu, Frederick and Kale, Satyen and Sundararajan, Mukund},
  journal={Advances in Neural Information Processing Systems},
  year={2020}
}

@article{hammoudeh2024training,
  title={Training data influence analysis and estimation: A survey},
  author={Hammoudeh, Zayd and Lowd, Daniel},
  journal={Machine Learning},
  volume={113},
  number={5},
  pages={2351--2403},
  year={2024},
  publisher={Springer}
}

@article{agarwal2017second,
  title={Second-order stochastic optimization for machine learning in linear time},
  author={Agarwal, Naman and Bullins, Brian and Hazan, Elad},
  journal={Journal of Machine Learning Research},
  year={2017},
}

@inproceedings{chhabra2024what,
title={What Data Benefits My Classifier? Enhancing Model Performance and Interpretability through Influence-Based Data Selection},
author={Chhabra, Anshuman and Li, Peizhao and Mohapatra, Prasant and Liu, Hongfu},
booktitle={International Conference on Learning Representations},
year={2024}
}

@article{kwon2023datainf,
  title={Datainf: Efficiently estimating data influence in lora-tuned llms and diffusion models},
  author={Kwon, Yongchan and Wu, Eric and Wu, Kevin and Zou, James},
  journal={arXiv preprint arXiv:2310.00902},
  year={2023}
}

@article{grosse2023studying,
  title={Studying large language model generalization with influence functions},
  author={Grosse, Roger and Bae, Juhan and Anil, Cem and Elhage, Nelson and Tamkin, Alex and Tajdini, Amirhossein and Steiner, Benoit and Li, Dustin and Durmus, Esin and Perez, Ethan and others},
  journal={arXiv preprint arXiv:2308.03296},
  year={2023}
}

@inproceedings{he2024large,
  title={Large-scale dataset pruning with dynamic uncertainty},
  author={He, Muyang and Yang, Shuo and Huang, Tiejun and Zhao, Bo},
  booktitle={IEEE/CVF Conference on Computer Vision and Pattern Recognition},
  year={2024}
}

@article{qin2023infobatch,
  title={Infobatch: Lossless training speed up by unbiased dynamic data pruning},
  author={Qin, Ziheng and Wang, Kai and Zheng, Zangwei and Gu, Jianyang and Peng, Xiangyu and Xu, Zhaopan and Zhou, Daquan and Shang, Lei and Sun, Baigui and Xie, Xuansong and others},
  journal={arXiv preprint arXiv:2303.04947},
  year={2023}
}

@article{toneva2018empirical,
  title={An empirical study of example forgetting during deep neural network learning},
  author={Toneva, Mariya and Sordoni, Alessandro and Combes, Remi Tachet des and Trischler, Adam and Bengio, Yoshua and Gordon, Geoffrey J},
  journal={arXiv preprint arXiv:1812.05159},
  year={2018}
}

@article{raju2021accelerating,
  title={Accelerating deep learning with dynamic data pruning},
  author={Raju, Ravi S and Daruwalla, Kyle and Lipasti, Mikko},
  journal={arXiv preprint arXiv:2111.12621},
  year={2021}
}

@inproceedings{wei2022learning,
  title={Learning with Noisy Labels Revisited: A Study Using Real-World Human Annotations},
  author={Wei, Jiaheng and Zhu, Zhaowei and Cheng, Hao and Liu, Tongliang and Niu, Gang and Liu, Yang},
  booktitle={International Conference on Learning Representations},
  year={2022}
}

@article{yang2025layer,
  title={Layer-Aware Influence for Online Data Valuation Estimation},
  author={Yang, Ziao and Huang, Longbo and Liu, Hongfu},
  journal={arXiv preprint arXiv:2510.16007},
  year={2025}
}

@inproceedings{he2016deep,
  title={Deep residual learning for image recognition},
  author={He, Kaiming and Zhang, Xiangyu and Ren, Shaoqing and Sun, Jian},
  booktitle={Computer Vision and Pattern Recognition},
  year={2016}
}

@article{evans2024data,
  title={Data curation via joint example selection further accelerates multimodal learning},
  author={Evans, Talfan and Parthasarathy, Nikhil and Merzic, Hamza and Henaff, Olivier},
  journal={Advances in Neural Information Processing Systems},
  year={2024}
}

@inproceedings{udandarao2025active,
  title={Active data curation effectively distills large-scale multimodal models},
  author={Udandarao, Vishaal and Parthasarathy, Nikhil and Naeem, Muhammad Ferjad and Evans, Talfan and Albanie, Samuel and Tombari, Federico and Xian, Yongqin and Tonioni, Alessio and H{\'e}naff, Olivier J},
  booktitle={Computer Vision and Pattern Recognition Conference},
  year={2025}
}

@inproceedings{song2019selfie,
  title={Selfie: Refurbishing unclean samples for robust deep learning},
  author={Song, Hwanjun and Kim, Minseok and Lee, Jae-Gil},
  booktitle={International Conference on Machine Learning},
  year={2019}
}

@inproceedings{bossard2014food,
  title={Food-101--mining discriminative components with random forests},
  author={Bossard, Lukas and Guillaumin, Matthieu and Van Gool, Luc},
  booktitle={European Conference on Computer Vision},
  pages={446--461},
  year={2014},
  organization={Springer}
}

\newpage
\appendix

{\large\textbf{Appendix}}
\section{Descriptions of Datasets and Models}
\label{app:data}

\paragraph{Datasets.}
We evaluate DIVE on six noisy-label image classification benchmarks, summarized in
Table~\ref{tab:datasets}. \textit{CIFAR-10N} and \textit{CIFAR-100N}~\citep{wei2022learning}
augment the original clean \textit{CIFAR-10/CIFAR-100} training images with human re-annotations
collected via Amazon Mechanical Turk. We use all three released label sets for CIFAR-10N
--- \textit{aggregate}, \textit{random}, and \textit{worst}, denoted \textit{CIFAR-10Na}, \textit{CIFAR-10Nr},
and \textit{CIFAR-10Nw} respectively --- corresponding to noise rates of approximately 9\%, 17\%, and
40\%, together with the single released label set for \textit{CIFAR-100N} (${\sim}40\%$ noise). Both
datasets contain 50,000 training images over 10 and 100 classes respectively, evaluated on
the standard 10,000-image \textit{CIFAR} test split. \textit{Animal-10N}~\citep{song2019selfie}
consists of 50,000 training and 5,000 test images spanning 10 visually confusable animal
classes (e.g., wolf/husky, hamster/guinea pig), human-annotation noise at approximately 8\%. \textit{Food-101N}~\citep{bossard2014food}
contains 75,750 training and 25,250 test images across 101 food categories, with labels
collected from web search results rather than human curation, carrying approximately 20\%
intrinsic label noise. Across the six benchmarks, noise rates range from ${\sim}8\%$ to
${\sim}40\%$.

For \textit{CIFAR-10N} and \textit{CIFAR-100N}, which retain clean pre-noise labels for the training split,
we construct the clean validation reference by holding out 1,000 clean-labeled samples from
the training set before noise is applied, and resample this pool at each scoring step. \textit{Animal-10N} and \textit{Food-101N} do not provide clean labels for their training data,
so we instead set aside a class-stratified 1/5 of each dataset's clean test split for
validation and report test accuracy on the remaining 4/5. 


\begin{table}[h]
\centering
\caption{Summary statistics of the six image benchmarks used in our experiments.}
\label{tab:datasets}
\begin{tabular}{lccccc}
\toprule
Dataset & \# Train & \# Test & \# Classes & Noise Type & Noise Rate \\
\midrule
\textit{CIFAR-10Na}  & 50{,}000 & 10{,}000 & 10  & Human-annotated (aggregate) & ${\sim}9\%$  \\
\textit{CIFAR-10Nr}  & 50{,}000 & 10{,}000 & 10  & Human-annotated (random)    & ${\sim}17\%$ \\
\textit{CIFAR-10Nw}  & 50{,}000 & 10{,}000 & 10  & Human-annotated (worst)     & ${\sim}40\%$ \\
\textit{CIFAR-100N}  & 50{,}000 & 10{,}000 & 100 & Human-annotated             & ${\sim}40\%$ \\
\textit{Animal-10N}  & 50{,}000 & 5{,}000  & 10  & Human-annotation       & ${\sim}8\%$  \\
\textit{Food-101N}   & 75{,}750 & 25{,}250 & 101 & Intrinsic web-label         & ${\sim}20\%$ \\
\bottomrule
\end{tabular}
\end{table}

\paragraph{Models.}
All image experiments use a ResNet-9~\citep{he2016deep} backbone, the compact
fast-training architecture popularized by David Page's DAWNBench ``How to Train Your
ResNet'' recipe. It consists of an initial $3\times3$ conv--batch-norm--ReLU stem (64
channels), followed by three downsampling stages at 128, 256, and 512 channels (each a
$3\times3$ conv--BN--ReLU block with $2\times2$ max-pooling), with two identity residual
blocks --- each two stacked conv--BN--ReLU layers with a skip connection --- inserted after
the 128-channel and 512-channel stages. A final \emph{global} max-pool (rather than a
fixed-window pool) collapses the spatial map to a single 512-dimensional feature vector per
image regardless of its size, making the backbone resolution-adaptive: the same 512-d output
applies uniformly across all six benchmarks despite their differing native resolutions
($32\times32$ for \textit{CIFAR-10N/100N}, $64\times64$ for \textit{Animal-10N}, $96\times96$ for Food-101N).
This backbone feeds into a three-layer MLP classification head
($512 \to 256 \to 128 \to \text{\#classes}$) with ReLU activations between layers. By default, sample-level influence scores are computed using
gradients of only the \emph{final} dense layer of the head, matching the paper's reported
default configuration (``last-1'').


\section{Additional Experimental Results}\label{app:experiment}
\label{app:indepth-extra}

\begin{figure}[h]
\centering
\includegraphics[width=\linewidth]{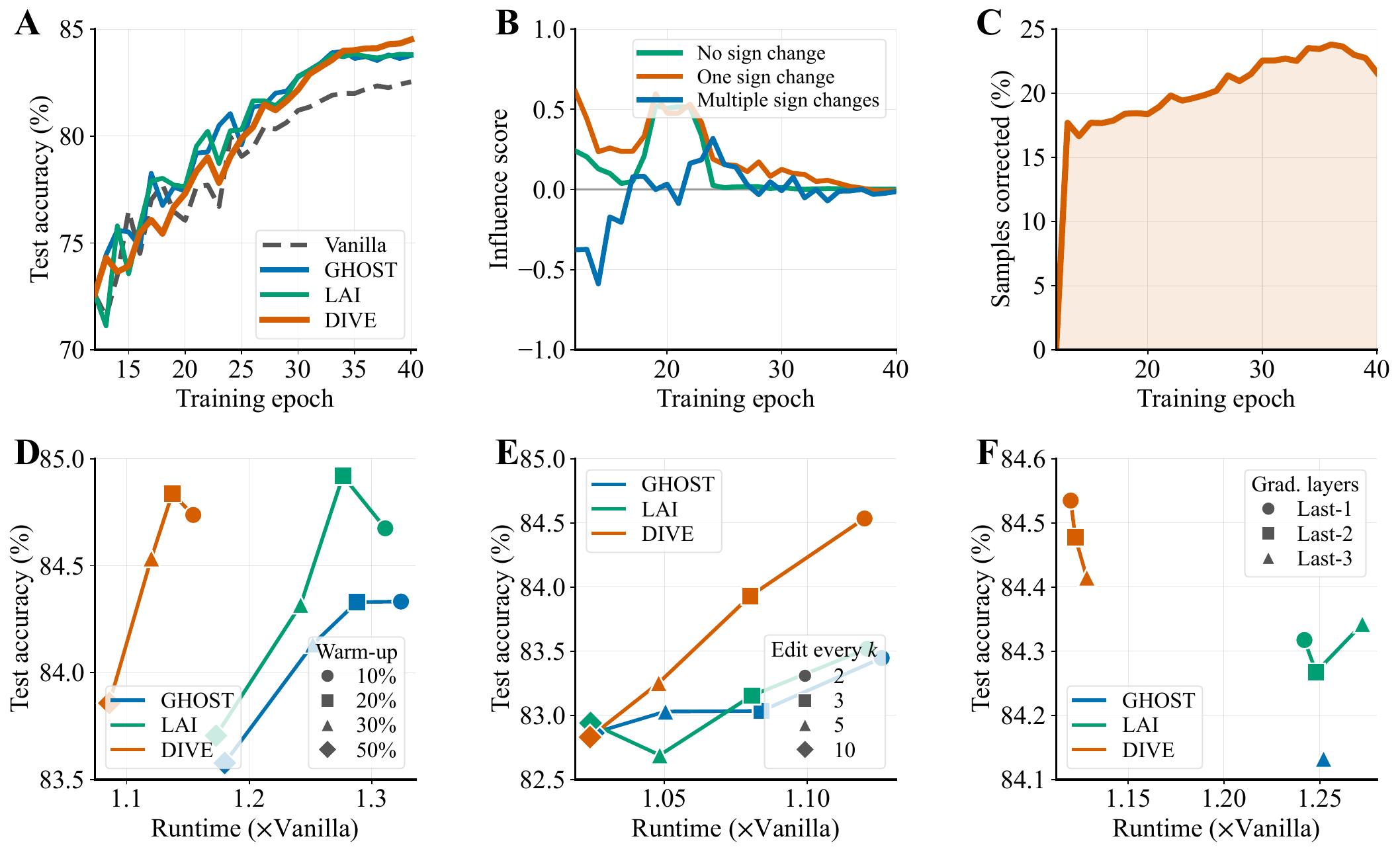}
\caption{In-depth exploration of DIVE on \textit{Animal-10N}.}\vspace{-4mm}
\label{fig:indepth_animal}
\end{figure}

\begin{figure}[h]
\centering
\includegraphics[width=\linewidth]{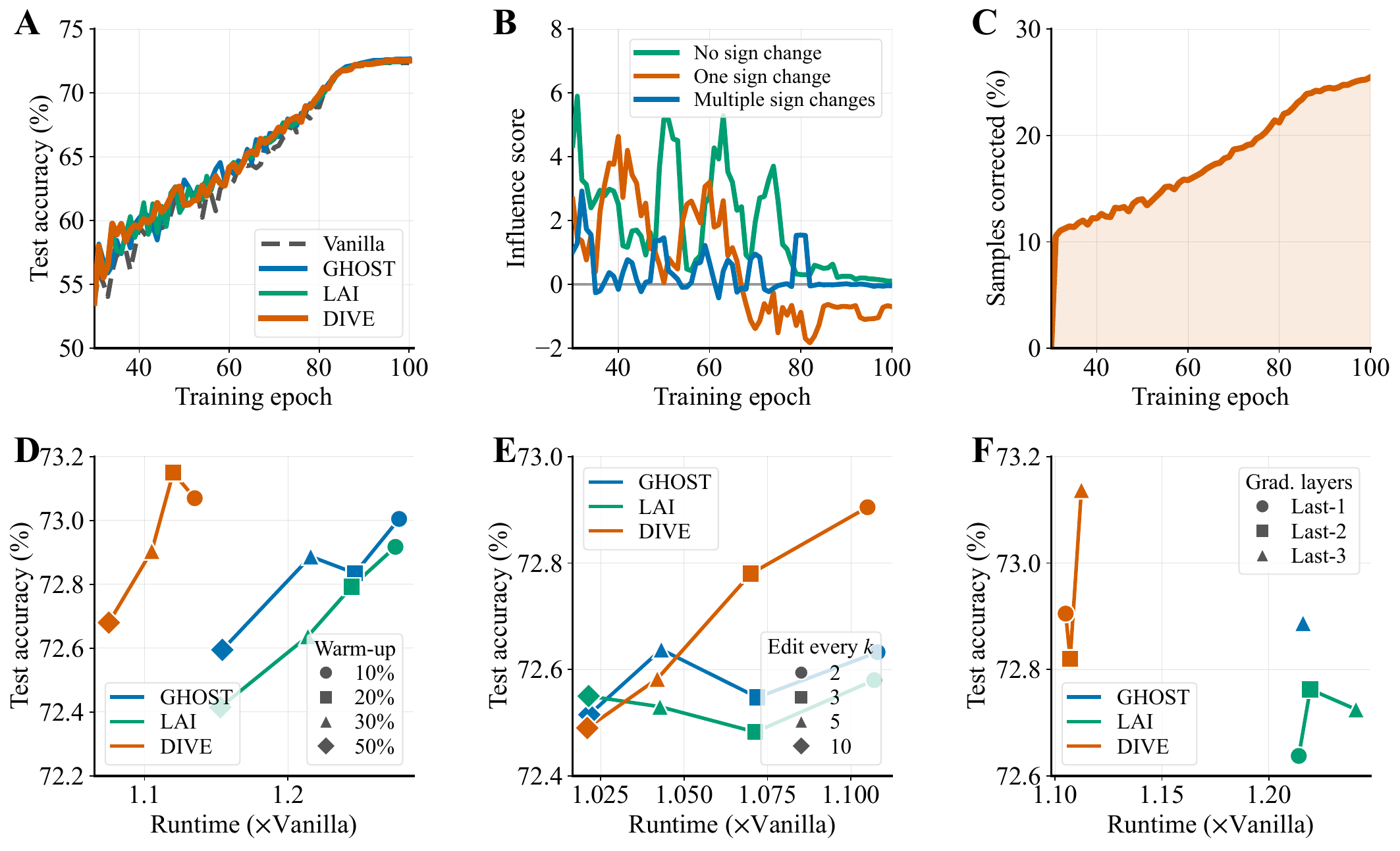}
\caption{In-depth exploration of DIVE on \textit{Food-101N}.}\vspace{-4mm}
\label{fig:indepth_food}
\end{figure}

The in-depth study in the main text uses \textit{CIFAR-10N-w}, whose labels are corrupted by synthetic noise. To confirm that the same behavior holds under real-world noise, we repeat the analysis on \textit{Animal-10N} and \textit{Food101-N}, shown in Figures~\ref{fig:indepth_animal} and~\ref{fig:indepth_food}. Both datasets tell the same story as \textit{CIFAR-10N-w}. After the warm-up phase, DIVE attains the highest test accuracy among the curation methods, and it keeps correcting a sizeable, time-varying fraction of the training set rather than a fixed subset. Tracking individual samples reveals the mechanism behind this: the sign of a sample's influence is not stable, so a sample that is beneficial at one stage of training can become detrimental later. The warm-up, editing-frequency, and scoring-layer ablations lead to the same conclusion as before, with DIVE reaching higher accuracy than GHOST and LAI at a lower training overhead. The main difference from \textit{CIFAR-10N-w} is that these datasets carry less noise, so the influence scores gradually shrink toward zero as the model converges and most of the sign changes occur in the earlier epochs.

\end{document}